\documentclass{article}

\pdftrailerid{}

\usepackage{colm2024_conference}

\usepackage{graphicx}
\usepackage{float}
\usepackage{placeins}
\usepackage{caption}
\usepackage[utf8]{inputenc}
\usepackage[T1]{fontenc}
\usepackage{booktabs}
\usepackage{algorithm}
\usepackage{algorithmic}
\usepackage{listings}
\usepackage{amsmath}
\usepackage{amssymb}
\usepackage{array}
\usepackage{colortbl}
\usepackage{multirow}
\usepackage{siunitx}
\usepackage{tabularx}
\usepackage[capitalise,noabbrev]{cleveref}

\newcommand{\system}{NavMCP}
\newcommand{\planner}{Qwen3.6-Plus}
\newcommand{\executor}{Qwen-RobotNav}
\newcommand{\gapname}{episodic interface gap}
\newcommand{\termAcc}{59.1}
\newcommand{\termDelta}{14.9}
\newcommand{\obsTermAcc}{68.1}
\newcommand{\obsTermDelta}{5.9}
\newcolumntype{Y}{>{\centering\arraybackslash}X}
\definecolor{TableHeader}{RGB}{237,231,247}
\definecolor{TableHighlight}{RGB}{246,241,251}
\definecolor{TableGroup}{RGB}{249,246,252}
\definecolor{TableRule}{RGB}{91,76,105}
\newcommand{\modernTableSetup}{%
    \setlength{\aboverulesep}{0pt}%
    \setlength{\belowrulesep}{0pt}%
    \setlength{\extrarowheight}{0pt}%
    \arrayrulecolor{TableRule}%
}
\lstdefinestyle{promptstyle}{%
    basicstyle=\ttfamily\scriptsize,
    backgroundcolor=\color{TableGroup},
    frame=single,
    rulecolor=\color{TableRule},
    breaklines=true,
    breakatwhitespace=false,
    columns=fullflexible,
    keepspaces=true,
    showstringspaces=false,
    aboveskip=5pt,
    belowskip=7pt
}
\title{Scaffolding Foundation Models into Physical-World Agents Pushes the Frontier of Long-Horizon Navigation}

\author{%
  \textbf{Zixing Lei$^{1,3,*}$, Gengze Zhou$^{4,*}$, Xiong-Hui Chen$^{2,*}$, Jiazhao Zhang$^{5}$, Yiyang Huang$^{2}$, Hang Yin$^{6}$, Haoqi Yuan$^{5}$, Qi Wu$^{4}$, Weixin Li$^{3}$, Siheng Chen$^{1}$} \\
  {\small
  $^1$Shanghai Jiao Tong University \quad $^2$Qwen Team, Alibaba Inc \quad \mbox{$^3$Zhongguancun Academy} \\
  $^4$AIML, Adelaide University \quad $^5$Peking University \quad $^6$Tsinghua University \\
  $^*$Equally contributed.}
}

\begin{document}

\maketitle

\begin{figure}[H]
\centering
\includegraphics[width=0.99\textwidth]{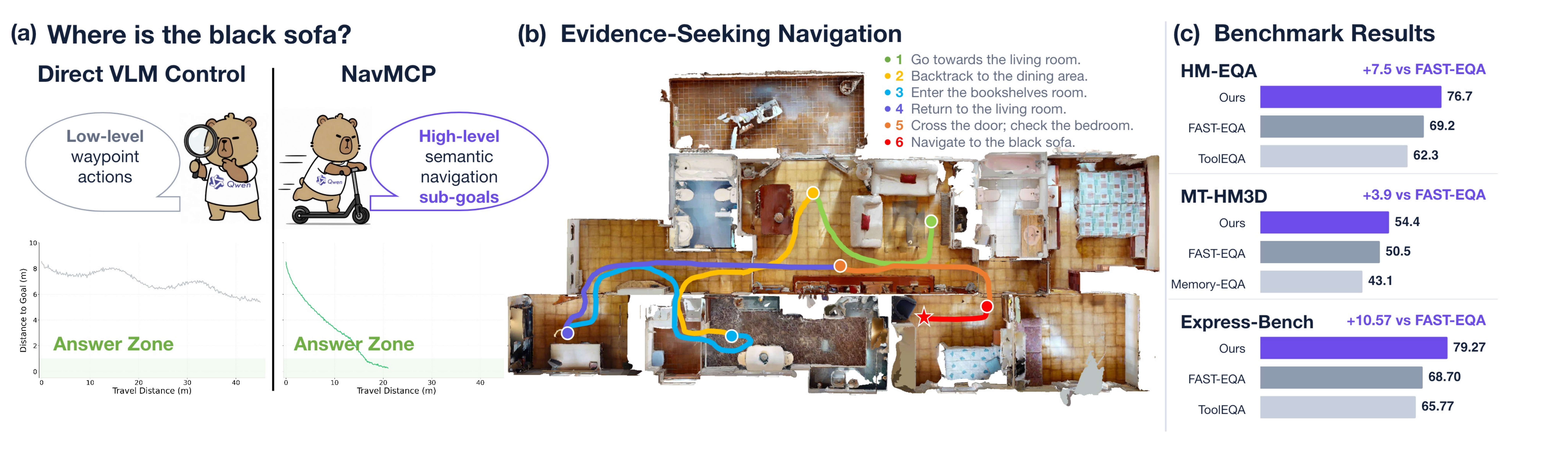}
\caption{(a) A VLM directly controlling short waypoint actions may struggle with long-horizon progress. Coupling it with an NFM lets the VLM issue semantic navigation sub-goals while the NFM performs closed-loop execution. (b) For an EQA question, the VLM converts evidence-seeking decisions into semantic navigation calls that search relevant rooms, rule out unlikely regions, and localize the target evidence. (c) Table~\ref{tab:main} values compare \system{} with FAST-EQA and the corresponding benchmark-specific baseline: ToolEQA on HM-EQA and Express-Bench, and Memory-EQA on MT-HM3D.}
\label{fig:teaser}
\end{figure}

\begin{abstract}
Long-horizon physical-world agents must reason over distant task goals while grounding each decision into reliable closed-loop behavior.
These capabilities are split across today's foundation models: vision-language models (VLMs) can infer missing information and adapt high-level plans, but repeatedly grounding their decisions into navigation actions remains brittle and inefficient, whereas navigation foundation models (NFMs) execute semantic goals robustly but typically operate as bounded episodes without persistent task-level reasoning.
We argue that these limitations are complementary and introduce \system{}, an agentic scaffolding framework that couples a VLM reasoning agent with an NFM executor for long-horizon physical-world exploration.
The VLM decides what evidence to seek, where to search, and when to stop, while the NFM grounds each semantic sub-goal into closed-loop navigation.
\system{} structures this collaboration through three channels: intent translates evidence needs into semantic navigation calls, observation converts complete rollouts into source-grounded trajectory evidence, and memory accumulates findings, negative evidence, and unresolved goals across calls.
Together, these channels turn isolated navigation rollouts into persistent embodied interaction without retraining either foundation model.
We instantiate \system{} on Embodied Question Answering (EQA), a demanding setting in which an agent must explore for visual evidence before answering situated questions.
\system{} achieves state-of-the-art performance on HM-EQA, MT-HM3D, and EXPRESS-Bench; under matched agent and executor backbones, it outperforms an episodic interface by \termDelta{} percentage points on HM-EQA.
On a Unitree Go2, \system{} reaches 78.3\% success, with its margin over the strongest baseline growing from 10 to 45 points as the task horizon increases.
These results demonstrate the potential of scaffolding complementary foundation models into long-horizon physical-world agents.
\end{abstract}

\section{Introduction}
\label{sec:intro}

{\looseness=-1
Long-horizon physical-world agency requires two capabilities that today's foundation models provide separately.
An agent must reason over distant goals and changing evidence while continuously grounding its decisions into reliable actions.
Vision-language models (VLMs) can infer what is missing, decompose goals, and adapt high-level plans, but repeatedly translating such reasoning into navigation actions becomes brittle and inefficient over long horizons.
Navigation foundation models (NFMs) are strong at the complementary problem: they map egocentric observations and semantic goals to closed-loop actions or waypoints, yet typically execute one bounded goal at a time without deciding what to seek next or retaining task state across episodes.
The central challenge is therefore not simply to build a stronger reasoner or navigator, but to organize their complementary capabilities into persistent physical-world agency.\par}

{\looseness=-1
Recent NFMs such as SAME, Uni-NaVid, NavFoM, ABot-N0/N1, and Qwen-RobotNav unify instruction following, object and point-goal navigation, point-of-interest search, and target following~\citep{zhou2025same,uninavid,navfom,chu2026abot,zhang2026qwenrobotnav,gong2026abotn1}.
Their strong navigation performance and zero-shot transfer make them powerful embodied executors, while VLM agents provide the open-ended reasoning needed to select and revise goals over a longer task horizon.
This complementarity suggests a natural division of labor: the VLM decides what evidence to seek, where to search, and when to stop, while the NFM determines how to ground each semantic sub-goal into closed-loop motion.
We call this paradigm \system{}: the NFM extends the VLM's action horizon, and the VLM agent extends the NFM's reasoning horizon.
Rather than asking either model to solve the full embodied task alone, \system{} scaffolds their specialized capabilities into a single long-horizon agent.\par}

{\looseness=-1
Realizing \system{} requires more than wrapping a navigator in a conventional tool call.
Under an \emph{episodic tool interface}, the agent submits one navigation instruction and receives only terminal status and a final observation, creating an \emph{\gapname{}} with multi-step evidence acquisition.
\textbf{Mismatch between instruction and intent.} A route-level instruction does not capture the agent's evidence need, search mode, budget, and constraints.
\textbf{Intermediate observation loss.} Useful evidence may appear anywhere along a rollout---an object passed in a hallway or a room glimpsed through a doorway---but a terminal-only return discards everything observed before the final pose.
\textbf{Lack of cross-call accumulation.} Independent calls do not retain searched regions, ruled-out hypotheses, uncertain findings, or negative outcomes.
Without addressing these gaps, an NFM remains an episodic tool rather than part of a long-horizon agent.\par}

{\looseness=-1
\system{} closes the \gapname{} with an evidence-centric scaffold organized into three channels (\Cref{fig:teaser}).
The intent channel translates an evidence need into a semantic navigation call with a mode, sub-goal, budget, and constraints, allowing the agent to specify what to seek without micromanaging low-level control.
The observation channel converts a complete rollout into source-grounded journey evidence containing keyframes, salient observations, uncertainty, and unexplored-region cues.
The memory channel carries checked regions, negative evidence, unresolved goals, and answer support across calls.
Together, these channels make NFM execution controllable, inspectable, and persistent: isolated navigation episodes become reusable embodied experience for subsequent reasoning and action.\par}

{\looseness=-1
We instantiate and evaluate \system{} on Embodied Question Answering (EQA), where an agent must explore a physical environment, acquire visual evidence, and answer a situated question~\citep{das2018embodied,gordon2018iqa}.
EQA provides a demanding test of long-horizon agency because distant or initially invisible evidence requires the agent to form search hypotheses, visit multiple regions, accumulate positive and negative observations, and decide when the evidence is sufficient to answer.
Across HM-EQA, MT-HM3D, and EXPRESS-Bench, \system{} improves over the strongest reported results; under matched reasoning and evaluation conditions, it leads FAST-EQA by 10.5 percentage points on HM-EQA.
With the agent and navigator fixed, replacing \system{} with an episodic interface costs \termDelta{} points, directly measuring the value of persistent scaffolding.
On a Unitree Go2, \system{} reaches 78.3\% success in real multi-room search, and its margin over the strongest baseline grows from 10 to 45 points as the task horizon increases.
These results show that scaffolding VLM reasoning and NFM execution is an effective route to long-horizon physical-world agents.\par}

Our contributions are:
\enlargethispage{\baselineskip}
\begin{enumerate}
    \item We introduce \system{}, an agentic scaffolding framework that organizes the complementary capabilities of a VLM reasoning agent and an NFM executor into a long-horizon physical-world agent.
    \item We develop an evidence-centric three-channel design that communicates navigation intent, grounds complete rollouts as reusable observations, and maintains task state across calls, thereby extending both the VLM's action horizon and the NFM's reasoning horizon.
    \item We validate \system{} on three simulated EQA benchmarks and a Unitree Go2, achieving state-of-the-art results across all three benchmarks and demonstrating gains that grow with the real-world task horizon.
\end{enumerate}

\section{Related Work}
\label{sec:related}

\paragraph{Embodied Question Answering.}
EQA, introduced by \citet{das2018embodied}, requires navigation to answer situated questions.
Recent systems improve exploration, memory, refinement, scene representation, and multi-agent search using frontiers~\citep{explore-eqa,memoryeqa}, memory guidance~\citep{lu2026memory}, goal-directed policies~\citep{fast-eqa}, scene graphs~\citep{grapheqa}, and distributed LLM agents~\citep{mele}; benchmarks now span open-vocabulary, free-form, and urban settings~\citep{openeqa,express-bench,cityeqa}.
\system{} instead uses an NFM as the physical layer while preserving trajectory evidence.

\paragraph{Vision-Language Navigation and Navigation Foundation Models.}
VLN has advanced from instruction-conditioned navigation~\citep{anderson2018vln} through augmentation, pretraining, transformers, video planning, and LLM reasoning~\citep{fried2018speaker,hao2020prevalent,chen2022think,hong2021vln-bert,zhang2024navid,navgpt,zhou2024navgpt}.
Semantic approaches add commonsense and language, belief, or frontier maps~\citep{esc_nav,vlmaps,zhou2026beliefmapnav,yokoyama2024vlfm}; navigation foundation models and generic policies instead unify tasks through vision-language-action interfaces~\citep{zhou2025same,uninavid,navfom,chu2026abot,zhang2026qwenrobotnav}.
On VLNVerse, Qwen-RobotNav-8B achieves 63.75\% SR and 57.93\% SPL under fine-grained instructions, and 46.59\% SR and 41.54\% SPL under coarse-grained instructions~\citep{zhang2026qwenrobotnav}.
\system{} targets the episodic tool interface, where one instruction produces one terminal return, and turns navigation rollouts into reusable evidence.

\paragraph{Tool-Augmented Embodied Agents and Context Management.}
LLMs plan embodied tasks through affordance grounding, feedback, code, planning, and skills~\citep{saycan,innermonologue,codeaspolicies,sayplan,voyager}.
Reasoning-acting and tool-use methods~\citep{react,reflexion,toolformer} extend to EQA and manipulation~\citep{tooleqa,lei2026towards}, while long-horizon agents use memory and prompt compression~\citep{memgpt,llmlingua}.
These systems support tool selection and context management but typically reduce each call to a terminal outcome; \system{} instead preserves source-grounded en-route and negative evidence across calls.

\section{Method}
\label{sec:method}
\subsection{System Formulation}
\label{sec:system_formulation}

We formulate EQA with a navigation foundation model as an agent--executor interface.
Given a question $q$ and an initial observation, a VLM agent must gather sufficient visual evidence to produce an answer $y$, deciding where to search and when to stop.
The agent maintains task hypotheses, evidence needs, and context, while a navigation executor handles one semantic sub-goal at a time.
We instantiate the executor with \executor{}~\citep{zhang2026qwenrobotnav}, which maps the agent's natural-language sub-goal, current egocentric observation, and navigation state to a short-horizon waypoint trajectory.
At turn $t$, the agent reads $C_t$ and either proposes $y_t$ or emits an evidence need $u_t$; in the latter case, the executor runs one rollout and returns trajectory-level observations.
This division makes repeated rollouts useful without retraining the navigator or asking it to answer directly: the executor supplies grounded experience, while the agent performs evidence reasoning and answer commitment.

\begin{figure*}[t]
\centering
\includegraphics[width=0.98\textwidth]{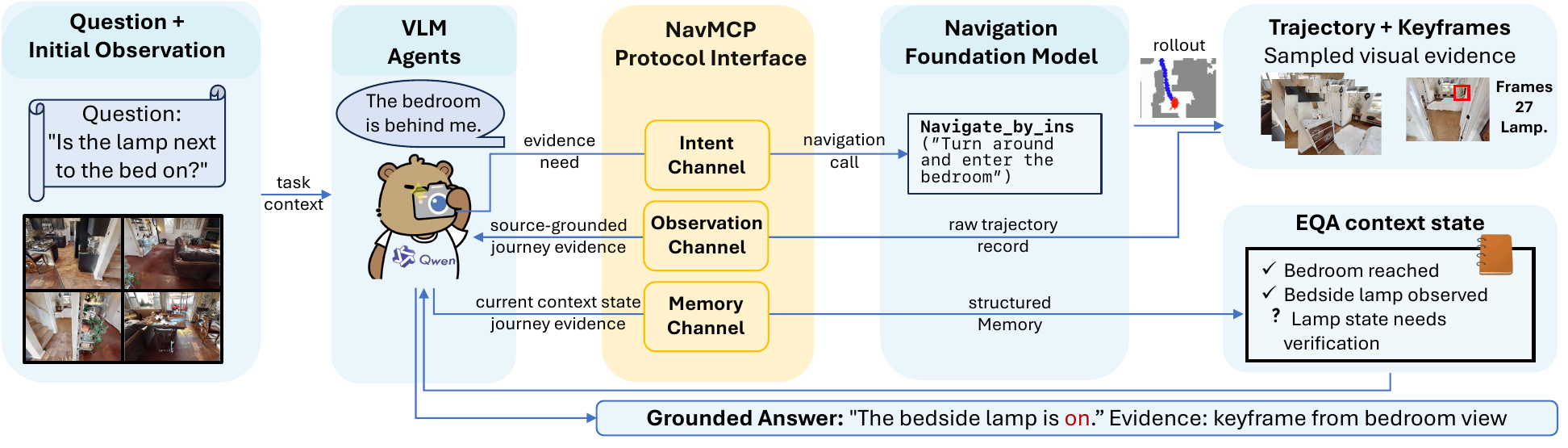}
\caption{\textbf{System overview of \system{}.} Given an EQA question and observation, a VLM agent uses \system{} to call a navigation foundation model as an embodied executor. The navigation foundation model receives a natural-language navigation sub-goal and the current egocentric observation and state, and outputs a short-horizon waypoint trajectory. The resulting trajectory observations are then converted back into evidence for the VLM agent.}
\label{fig:system_overview}
\end{figure*}

The \gapname{} defined in \Cref{sec:intro} structures this interface.
The agent must communicate evidence-seeking intent without low-level control, the executor's rollout must be converted into evidence useful for answering rather than reduced to terminal feedback, and salient evidence must persist across repeated calls.
\system{} addresses these requirements with three channels: intent, observation, and memory, each closing one mismatch.
\Cref{fig:system_overview} illustrates the end-to-end system, and \Cref{fig:navmcp_channels} expands the three scaffold channels.
See details in \Cref{alg:agent_loop} of the appendix.

\subsection{Intent Channel: From Evidence Needs to Navigation Calls}
\label{sec:nfm_tool}

The intent channel turns an agent's evidence need into one navigation episode, closing the mismatch between instruction and intent.
This channel is needed because the agent reasons about missing information, while the navigation foundation model acts through embodied rollouts and consumes route-level instructions.
The agent should therefore specify where or what to search for, but it should not issue simulator actions or micromanage the path.
At turn $t$, the agent emits a semantic navigation call that specifies a mode, a natural-language sub-goal, a step budget, and optional constraints; the schema is given in \Cref{sec:appendix_protocol_details}.
The key abstraction is that the sub-goal and constraints specify intent rather than controls.
For example, the agent can request ``search the kitchen counters for a coffee maker'' rather than prescribe actions.

The executor adapter turns each intent call into an instruction-following episode for the navigation foundation model.
We expose two modes that match the two common operating regimes used to train and evaluate navigation foundation models.
\texttt{navigate\_to\_object} corresponds to Object Navigation, where the instruction names an object or category to find.
\texttt{navigate\_by\_instruction} corresponds to Vision-and-Language Navigation, where the instruction describes a route, region, or scene-level goal.
This mode split gives the agent a simple way to choose between object-centric search and broader language-guided movement while keeping both modes backed by the same navigation executor.
The executor observes only egocentric RGB views, its pose, and an online map built from the explored trajectory, so the interface does not assume privileged access to unseen geometry or object locations.
Because the agent only depends on the call-and-return schema, the navigation model can be replaced by another executor if it supports the same intent modes and return format, enabling the executor comparisons in our experiments.

\subsection{Observation Channel: From Trajectories to Evidence}
\label{sec:nav_summary}

The observation channel translates an embodied rollout into agent-facing evidence, closing intermediate observation loss.
A terminal success flag or final pose is too weak for evidence-seeking tasks.
Useful observations may occur anywhere along the route, including objects passed en route, rooms that were partially searched, and exits that remain unexplored.
For each navigation call, the executor returns a raw trajectory record containing rollout status, path information, final pose, and sampled keyframes; the detailed record format is given in \Cref{sec:appendix_protocol_details}.
We sample keyframes along the trajectory rather than only at the final pose because answer evidence can appear before the executor stops.
The observation channel then uses a VLM-based trajectory summarizer to produce a grounded journey artifact covering sub-goal status, route context, observed objects, planning cues, and uncertainty.
The artifact separates evidence for the current decision from cues for future navigation.
Rooms and objects support answer reasoning, while spatial cues and unexplored exits help the agent decide where to search next.

The journey artifact is source-grounded to reduce unsupported trajectory claims.
Each object or room mention is tied to a source keyframe and confidence label.
The summarizer is instructed to report only visually observed evidence, mark uncertainty explicitly, and avoid turning absence from a single view into a definitive negative claim.
This design preserves intermediate observations lost by standard navigation feedback and returns grounded evidence to memory.

\begin{figure}[t]
\centering
\includegraphics[width=0.65\linewidth]{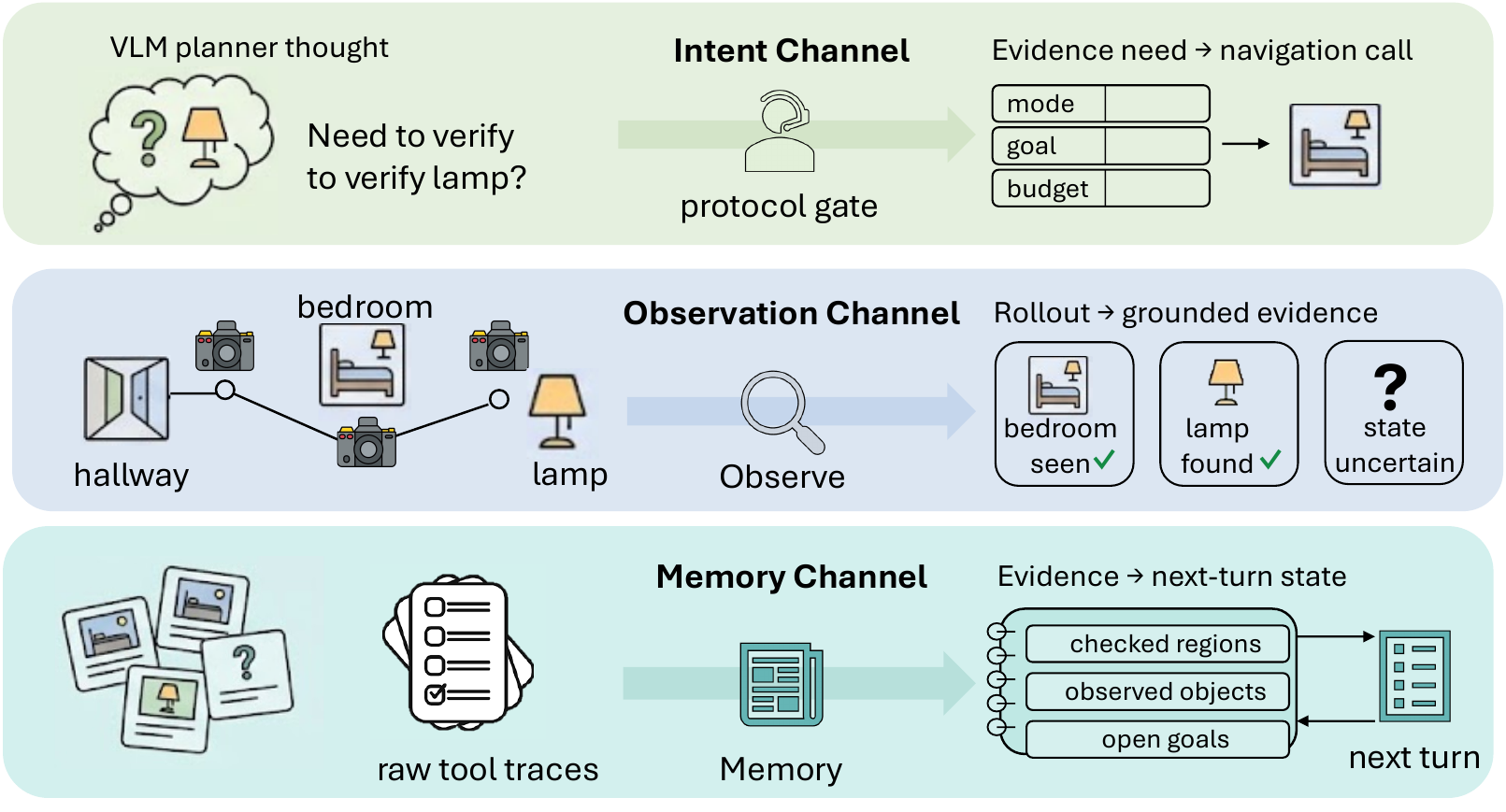}
\caption{\textbf{Three-channel \system{} interface.} Each channel closes one mismatch of the \gapname{}: the intent channel resolves the mismatch between instruction and intent by mapping the agent's evidence need into a structured navigation call, the observation channel prevents intermediate observation loss by converting the executor rollout into grounded trajectory evidence, and the memory channel enables cross-call accumulation by writing the evidence and remaining open goals into the next-turn EQA context state. These channels separate answer reasoning from low-level navigation while retaining visual evidence for later decisions.}
\label{fig:navmcp_channels}
\end{figure}

\subsection{Memory Channel: From Tool Returns to EQA Context State}
\label{sec:evidence_management}

The memory channel turns trajectory evidence into a compact EQA state for later reasoning turns, closing the lack of cross-call accumulation.
Without this state, the agent must repeatedly infer what has already been searched from long raw tool traces.
This is costly and brittle when earlier tool outputs are shortened under context limits.
We represent the EQA context state as $C_t = (H_t, E_t, U_t)$,
where $H_t$ is a compact interaction history, $E_t$ is a source-grounded evidence ledger, and $U_t$ stores unresolved goals for future planning.
After each journey artifact $z_t$ or auxiliary tool return, the agent updates $E_t$ and $U_t$ before the raw return can be compressed.
The ledger stores positive observations, searched areas, uncertain findings, negative evidence, and support for candidate decisions, while the unresolved-goal state tracks missing evidence for later intent calls.
Detailed ledger fields and update rules are given in \Cref{sec:appendix_protocol_details}.

Context maintenance follows a conservative compression policy.
Raw tool returns are compressed only after salient information has been externalized into $E_t$ or $U_t$ with source references, so long contexts can be reduced around persistent evidence rather than raw traces.
This policy makes compression safer because later reasoning depends on source-grounded ledger entries rather than on the full raw trace.

\paragraph{EQA instantiation.}
For EQA, \system{} implements $E_t$ as an evidence ledger for answering a question $q$ with a grounded response $y$.
The agent may answer only when the response is supported by the current observation, a journey artifact, a reviewed keyframe, or a notebook entry.
Positive answers should cite what was observed and where it was observed.
Negative answers should be backed by searched-area records covering likely locations for the target.
The agent can request local verification through auxiliary perception tools such as panoramic inspection, open-vocabulary detection, close-up object inspection, and saved-keyframe review.
These tools refine or verify ledger entries in our EQA instantiation; the protocol itself only requires source-grounded evidence updates.
Reported results therefore measure cross-call evidence management within an episode rather than memorization of test environments across episodes.

\section{Experiments}
\label{sec:experiment}
\subsection{Experimental Setup}
\label{sec:setup}

\paragraph{Evaluation questions.}
We organize the experiments around four questions, addressed in the following four subsections.
\textbf{(Q1)} Does \system{} improve overall EQA performance?
\textbf{(Q2)} Do the VLM agent and NFM executor complement each other?
\textbf{(Q3)} How does \system{} close the episodic interface gap?
\textbf{(Q4)} Does \system{} transfer to real robots as task horizons grow?

\paragraph{Benchmarks.}
We evaluate on three EQA benchmarks that stress complementary aspects of evidence gathering.
HM-EQA~\citep{explore-eqa} contains 500 multiple-choice questions across 267 HM3D scenes and evaluates single-question embodied search.
MT-HM3D~\citep{memoryeqa} introduces multi-target questions that require cross-room comparison and reasoning over observations collected in multiple regions.
EXPRESS-Bench~\citep{express-bench} contains 2,044 free-form questions and reports answer quality via LLM Score and exploration path quality via $E_\text{path}$.

\paragraph{Baselines.}
We compare against EQA systems that instantiate different interfaces between reasoning and navigation: Explore-EQA~\citep{explore-eqa} and Memory-EQA~\citep{memoryeqa} use exploration and memory mechanisms, Graph-EQA~\citep{grapheqa} uses structured scene representation, ToolEQA~\citep{tooleqa} augments reasoning with callable tools, FAST-EQA~\citep{fast-eqa} uses faster goal-directed exploration, and Fine-EQA~\citep{express-bench} is the reported baseline for EXPRESS-Bench.
For causal analysis, we replace the navigation foundation model with Random Walk or Frontier Exploration and use reactive, single-round, and no-agent variants.

\begin{table*}[t]
\centering
\small
\modernTableSetup
\renewcommand{\arraystretch}{1.16}
\setlength{\tabcolsep}{3.5pt}
\caption{\textbf{Cross-benchmark EQA performance.} Acc. denotes answer accuracy (\%). Norm. Agent Steps$^\dagger$ is the normalized high-level budget fraction; the equivalent-step compensation in \Cref{eq:main_equivalent_steps} applies only to methods that invoke an NFM, while non-NFM baselines retain their reported step accounting. Baselines are reported results, with $^*$ marking results reproduced by FAST-EQA. Our results are mean $\pm$ standard deviation over three independent runs. Cross-paper comparisons are contextual only.}
\label{tab:main}
\renewcommand{\tabularxcolumn}[1]{m{#1}}
\begin{tabularx}{\textwidth}{>{\raggedright\arraybackslash}X
    S[table-format=2.1,table-column-width=0.10\textwidth]
    S[table-format=1.2,table-column-width=0.10\textwidth]
    S[table-format=2.1,table-column-width=0.10\textwidth]
    S[table-format=1.2,table-column-width=0.10\textwidth]
    S[table-format=2.2,table-column-width=0.10\textwidth]
    S[table-format=2.2,table-column-width=0.10\textwidth]}
\toprule
\rowcolor{TableHeader}
\textbf{Method}
& \multicolumn{6}{c}{%
    \begin{tabular}{@{}
        >{\centering\arraybackslash}p{0.10\textwidth}
        >{\centering\arraybackslash}p{0.10\textwidth}
        >{\centering\arraybackslash}p{0.10\textwidth}
        >{\centering\arraybackslash}p{0.10\textwidth}
        >{\centering\arraybackslash}p{0.10\textwidth}
        >{\centering\arraybackslash}p{0.10\textwidth}@{}}
    \multicolumn{2}{c}{\textbf{HM-EQA}}
    & \multicolumn{2}{c}{\textbf{MT-HM3D}}
    & \multicolumn{2}{c}{\textbf{EXPRESS-Bench}} \\
    \cmidrule(lr){1-2}
    \cmidrule(lr){3-4}
    \cmidrule(lr){5-6}
    {\shortstack{\rule{0pt}{2.2ex}Acc.\\($\uparrow$)}}
    & {\shortstack{\rule{0pt}{2.2ex}Norm. Agent\\Steps$^\dagger$ ($\downarrow$)}}
    & {\shortstack{\rule{0pt}{2.2ex}Acc.\\($\uparrow$)}}
    & {\shortstack{\rule{0pt}{2.2ex}Norm. Agent\\Steps$^\dagger$ ($\downarrow$)}}
    & {\shortstack{\rule{0pt}{2.2ex}LLM Score\\($\uparrow$)}}
    & {\shortstack{\rule{0pt}{2.2ex}$E_{\text{path}}$\\($\uparrow$)}}
    \end{tabular}} \\
\midrule
Explore-EQA~\citep{explore-eqa}
& 58.4 & 0.52 & \multicolumn{1}{c}{\phantom{$^*$}36.2$^*$} & 0.64 & {--} & {--} \\
Graph-EQA~\citep{grapheqa}
& 63.5 & 0.20 & \multicolumn{1}{c}{\phantom{$^*$}45.6$^*$} & 0.45 & {--} & {--} \\
Memory-EQA~\citep{memoryeqa}
& 61.4 & 0.40 & 43.1 & 0.41 & {--} & {--} \\
Fine-EQA~\citep{express-bench}
& 56.0 & 0.54 & {--} & {--} & 63.95 & 25.58 \\
3D-Mem~\citep{yang20253d}
& 50.4 & 0.63 & {--} & {--} & {--} & {--} \\
ToolEQA~\citep{tooleqa}
& 62.3 & {--} & {--} & {--} & 65.77 & 25.82 \\
FAST-EQA~\citep{fast-eqa}
& 69.2 & 0.65 & 50.5 & 0.52 & 68.7 & 29.25 \\
\midrule
\rowcolor{TableHighlight}
\system{} (Ours)
& \multicolumn{1}{c}{\textbf{76.7$\pm$0.1}}
& \multicolumn{1}{c}{\textbf{0.15$\pm$0.01}}
& \multicolumn{1}{c}{\textbf{54.4$\pm$1.5}}
& \multicolumn{1}{c}{\textbf{0.19$\pm$0.01}}
& \multicolumn{1}{c}{\textbf{79.27$\pm$0.44}}
& \multicolumn{1}{c}{\textbf{33.96$\pm$0.75}} \\
\bottomrule
\end{tabularx}
\end{table*}

\paragraph{Metrics and step accounting.}
Answer quality is primary: accuracy (Acc.) is the percentage of questions answered correctly.
On EXPRESS-Bench, LLM Score rates free-form semantic correctness from 0--100, while $E_\text{path}$ weights it by the ratio of the reference sufficient path length to the traveled distance, rewarding accurate and efficient exploration.
For HM-EQA and MT-HM3D, we additionally report normalized equivalent agent steps.
In our setup, outer-agent VLM reasoning and API latency dominate wall-clock time, so we use the number of outer-agent VLM invocations, $H$, as the primary efficiency cost.
Under the direct-VLM-control protocol used by prior frontier baselines, one VLM decision can displace the agent by at most 3\,m; thus, 3\,m is the natural conversion unit rather than an arbitrary threshold.
For methods that invoke an NFM, each outer-agent invocation counts as one base step, and each NFM rollout incurs an additional charge for every extra 3\,m segment:
\begin{equation}
n_{\mathrm{eq}} = H + \sum_{k=1}^{K}\max\!\left(0,\left\lceil\frac{L_k}{3\,\mathrm{m}}\right\rceil-1\right),
\hat{n}_{\mathrm{eq}} = \frac{n_{\mathrm{eq}}}{N}.
\label{eq:main_equivalent_steps}
\end{equation}
Here $K$ is the number of NFM rollouts, $L_k$ is the path length of NFM rollout $k$, and $N$ is the scene-specific step budget.
The compensation term is not applied to non-NFM baselines, whose reported normalized steps follow their original direct-control accounting.
The metric measures high-level interaction and exploration rather than end-to-end computation; the full derivation and component-level accounting are given in \Cref{sec:appendix_steps}.

\paragraph{Implementation and evaluation protocol.}
The main results use \planner{} as the agent and \executor{}~\citep{zhang2026qwenrobotnav} as the navigation foundation model executor.
For controlled ablations, we use a fixed open-source Qwen3.5-397B-A17B agent unless noted otherwise and match episodes, initial states, budgets, and perception settings, so each comparison isolates the factor specified in its table caption.
Cross-paper results in \Cref{tab:main} are provided for context only; all causal claims rely on these controlled comparisons.
Further implementation details and evaluation statistics are provided in \Cref{sec:appendix_impl,sec:appendix_statistics}.

\subsection{Does \system{} Improve Overall EQA Performance?}
\label{sec:main_results}

Following FAST-EQA~\citep{fast-eqa} and common practice in prior EQA work, \Cref{tab:main} uses each baseline's latest reported result or its FAST-EQA reproduction when available. On HM-EQA, \system{} reaches 76.7\% accuracy, 7.5 percentage points above FAST-EQA. On MT-HM3D, it reaches 54.4\%, a 3.9-point improvement. On~EXPRESS-Bench, \system{} obtains an LLM Score of 79.27 and an $E_\text{path}$ of 33.96, improving over FAST-EQA by 10.57 and 4.71 points. Crucially, these gains come with substantially better efficiency: after compensating only the NFM rollouts for distance beyond the direct-control 3\,m unit, \system{} uses the lowest normalized agent-step fractions among reporting methods on HM-EQA and MT-HM3D (0.15 and 0.19), where outer-agent VLM calls dominate cost, and achieves the highest path efficiency on EXPRESS-Bench. Together, these results establish new state-of-the-art performance across all three EQA formats and show that coupling a reasoning VLM with a navigation foundation model is an effective and efficient paradigm for long-horizon EQA exploration. Because limited API and code availability prevents full alignment across papers, we next test the system-level advantage under matched conditions before characterizing agent, executor, and protocol contributions in \Cref{sec:ablation_nav,sec:analysis}.

\begin{table}[t]
\centering
\footnotesize
\modernTableSetup
\setlength{\extrarowheight}{2pt}
\renewcommand{\arraystretch}{1.14}
\caption{\textbf{Matched full-system comparison on HM-EQA.} All methods use the same Qwen3.5-397B-A17B agent, episodes, initial states, and budget; ToolEQA additionally uses the same perception toolset as \system{}. Methods retain their own pipelines and navigation interfaces. Values are mean $\pm$ standard deviation over three runs.}
\label{tab:matched_systems}
\setlength{\tabcolsep}{2.5pt}
\begin{tabularx}{\columnwidth}{X c}
\toprule
\rowcolor{TableHeader}
\textbf{Configuration} & \textbf{Acc. (\%)} \\
\midrule
Explore-EQA~\citep{explore-eqa} (re-evaluated) & 57.6 $\pm$ \textit{0.7} \\
ToolEQA~\citep{tooleqa} (our reproduction) & 60.8 $\pm$ \textit{0.9} \\
FAST-EQA~\citep{fast-eqa} (our reproduction) & 63.5 $\pm$ \textit{1.4} \\
\rowcolor{TableHighlight}
\system{} + \executor{}-8B & \textbf{74.0 $\pm$ \textit{0.3}} \\
\bottomrule
\end{tabularx}
\end{table}

\textbf{Matched full-system comparison.}
We use the same Qwen3.5 agent, episodes, initial states, and budget for all methods; ToolEQA also receives the same perception toolset as \system{}. As shown in \Cref{tab:matched_systems}, Explore-EQA, ToolEQA, and FAST-EQA obtain 57.6\%, 60.8\%, and 63.5\%, whereas \system{} with \executor{}-8B reaches 74.0\%, leading by 16.4, 13.2, and 10.5 points. \system{} therefore remains strongest after aligning the high-level reasoning and evaluation conditions.

\subsection{Do the VLM Agent and NFM Executor Complement Each Other?}
\label{sec:ablation_nav}

\begin{table}[t]
\centering
\footnotesize
\modernTableSetup
\setlength{\extrarowheight}{1pt}
\renewcommand{\arraystretch}{1.08}
\caption{\textbf{VLM--NFM architecture ablation.} Top: \executor{}-8B is fixed while the upper layer changes. Bottom: the full Qwen3.5 agent and \system{} harness are fixed while the executor changes. Executor variants are mean $\pm$ standard deviation over three runs.}
\label{tab:capability_sides}
\setlength{\tabcolsep}{2.5pt}
\begin{tabularx}{\columnwidth}{X c}
\toprule
\rowcolor{TableHeader}
\textbf{Configuration} & \textbf{Acc. (\%)} \\
\midrule
\rowcolor{TableGroup}
\multicolumn{2}{l}{\textit{Upper-level agent/harness capability (\executor{}-8B fixed)}} \\
No Agent (VLM from initial view) & 38.2 \\
Single-Round Agent (one panorama) & 58.4 \\
Reactive Agent (fixed explore loop) & 62.0 \\
Full Agent (Qwen3.5-397B-A17B) & 74.0 \\
\rowcolor{TableHighlight}
Full Agent (Qwen3.6-Plus) & \textbf{76.7} \\
\addlinespace[2pt]
\rowcolor{TableGroup}
\multicolumn{2}{l}{\textit{Lower-level navigation capability (full Qwen3.5 agent fixed)}} \\
\system{} + Random Walk & 60.9 $\pm$ \textit{3.1} \\
\system{} + Frontier Exploration & 65.3 $\pm$ \textit{0.2} \\
\system{} + StreamVLN~\citep{wei2026streamvln} & 69.3 $\pm$ \textit{0.4} \\
\system{} + \executor{}-4B & 73.3 $\pm$ \textit{0.3} \\
\rowcolor{TableHighlight}
\system{} + \executor{}-8B & \textbf{74.0 $\pm$ \textit{0.3}} \\
\bottomrule
\end{tabularx}
\end{table}

\Cref{tab:capability_sides} uses two controlled sweeps to show that the upper-level VLM agent and lower-level NFM executor complement each other in the two-level architecture.
\textbf{EQA performance improves substantially with stronger agent harnesses and reasoning models.}
With \executor{}-8B fixed, accuracy rises from 38.2\% without an agent through single-round (58.4\%) and reactive (62.0\%) variants to 74.0\% with the full Qwen3.5 harness; Qwen3.6-Plus further reaches 76.7\%.
These gains show that a capable NFM alone is insufficient without an upper-level agent that can adaptively plan, accumulate, and reuse evidence.
\textbf{EQA performance also improves substantially with a stronger navigation executor.}
With the full Qwen3.5 agent and \system{} harness fixed, accuracy rises from Random Walk (60.9\%) and Frontier Exploration (65.3\%) to StreamVLN (69.3\%), \executor{}-4B (73.3\%), and \executor{}-8B (74.0\%).
Thus, a capable upper layer cannot compensate for weak execution: stronger language-conditioned navigation converts the same evidence needs into more useful trajectories and observations.
Together, the two sweeps show that the VLM agent and NFM executor are complementary rather than substitutable, and end-to-end performance improves as either side becomes stronger.

\begin{figure*}[t]
\centering
\includegraphics[width=0.98\textwidth]{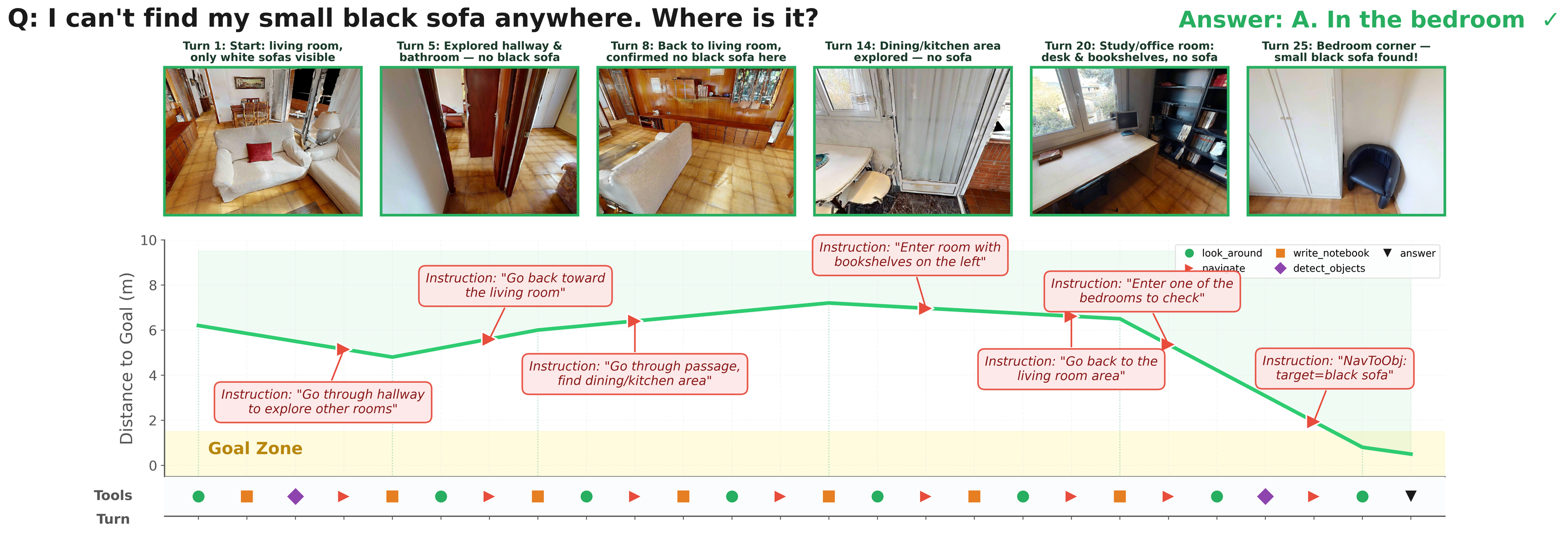}
\caption{\textbf{Evidence-guided search in an Explore-EQA episode} (find a small black sofa). Top: selected first-person observations; the agent records negative evidence in five rooms before finding the target in a bedroom corner. Middle: distance to goal across navigation calls (red triangles: instructions sent to the executor); the trajectory is deliberately non-monotonic, moving away from the goal to rule out plausible rooms before an object-targeted final call. Bottom: tool-call timeline over 25 turns.}
\label{fig:progressive_discovery}
\end{figure*}

\Cref{fig:progressive_discovery} illustrates this division of labor.
To find a small black sofa, the upper layer searches five rooms, records negative evidence to avoid repeats, and switches to an object-targeted request after narrowing the hypothesis.
The NFM executes each request and returns en-route observations.
The non-monotonic curve reflects evidence-driven exploration: the system moves away to eliminate plausible regions before converging on answer evidence.

\subsection{How Does \system{} Close the Episodic Interface Gap?}
\label{sec:analysis}

\begin{table}[t]
\centering
\small
\modernTableSetup
\setlength{\extrarowheight}{1pt}
\renewcommand{\arraystretch}{1.10}
\caption{\textbf{Protocol ablation on HM-EQA.} All rows fix the Qwen3.5-397B-A17B agent and \executor{}-8B executor. The episodic-interface baseline replaces the complete three-channel protocol; the observation-only terminal-return variant and remaining rows remove or weaken one channel or auxiliary capability.}
\label{tab:ablation_internal}
\begin{tabularx}{\columnwidth}{X
    S[table-format=2.1,table-column-width=0.18\columnwidth]
    S[table-format=-1.1,table-column-width=0.13\columnwidth]}
\toprule
\rowcolor{TableHeader}
\textbf{Configuration} & {\textbf{Acc.\ (\%)}} & {\textbf{$\Delta$}} \\
\midrule
\rowcolor{TableHighlight}
Full system & \multicolumn{1}{c}{\textbf{74.0}} & {--} \\
\addlinespace[2pt]
\rowcolor{TableGroup}
\multicolumn{3}{l}{\textit{Full episodic interface}} \\
\quad Episodic interface & \multicolumn{1}{c}{\termAcc} & \multicolumn{1}{c}{$-$\termDelta} \\
\addlinespace[2pt]
\rowcolor{TableGroup}
\multicolumn{3}{l}{\textit{Intent channel / navigation interface}} \\
\quad Single navigation mode & 72.0 & -2.0 \\
\addlinespace[2pt]
\rowcolor{TableGroup}
\multicolumn{3}{l}{\textit{Observation channel}} \\
\quad Terminal-only observation return & \multicolumn{1}{c}{\obsTermAcc} & \multicolumn{1}{c}{$-$\obsTermDelta} \\
\quad w/o Journey Analysis & 69.6 & -4.4 \\
\quad Sparse keyframes (8-step / max 8) & 71.2 & -2.8 \\
\addlinespace[2pt]
\rowcolor{TableGroup}
\multicolumn{3}{l}{\textit{Memory channel}} \\
\quad w/o EQA context state & 69.4 & -4.6 \\
\rowcolor{TableGroup}
\multicolumn{3}{l}{\textit{Auxiliary capabilities}} \\
\quad w/o Context Compaction & 73.1 & -0.9 \\
\quad w/o \texttt{zoom\_in\_object} & 73.4 & -0.6 \\
\bottomrule
\end{tabularx}
\end{table}

With both architectural layers fixed, \textbf{the complete episodic interface is insufficient.}
Replacing the three-channel protocol with an episodic interface reduces accuracy from 74.0\% to \termAcc\% ($-$\termDelta{} points), measuring the combined cost of the intent, observation, and cross-call accumulation gaps without changing the agent or executor backbone.

\textbf{Core protocol channels contribute differently.}
Restricting the intent channel to one navigation mode costs 2.0 points, showing that instruction-following and object-targeted calls support complementary searches.
For observation, changing only the observation return to terminal status and the final view costs \obsTermDelta{} points, removing journey analysis costs 4.4 points, and sparsifying keyframes costs 2.8 points, showing that rollout evidence must be captured and structured.
For memory, removing the EQA context state costs 4.6 points, showing that evidence and unresolved goals must persist across calls.
Observation and memory produce the two largest component-level effects.
Beyond the core protocol channels, auxiliary capabilities provide incremental gains.
Context compaction and the \texttt{zoom\_in\_object} verification tool improve accuracy by 0.9 and 0.6 points, respectively, suggesting that the harness can further benefit from additional context-management and verification capabilities.
Overall, intent specifies what to seek, observation preserves what the rollout reveals, and memory carries that evidence into subsequent decisions.

\subsection{Does \system{} Transfer to Real Robots as Task Horizons Grow?}
\label{sec:real_world}

\paragraph{Setup.}
We test the same agent--NFM interface beyond simulation under sensor noise, odometry drift, local recovery, and long routes.
Episodes use egocentric RGB, pose estimates, and an online map without privileged object locations or unseen geometry.
Six task types with ten episodes each span single-room, cross-room, and over-20\,m exploration; definitions are given in \Cref{sec:appendix_real_robot}.

\paragraph{Baselines.}
All three methods use the same \planner{} agent.
The reactive baseline retains \executor{} but removes multi-step deliberation and evidence accumulation; the frontier baseline replaces \executor{} with heuristic coverage while keeping the agent fixed.
These comparisons isolate orchestration and learned execution under real-world noise without repeating the full simulation ablations.

\begin{table}[t]
\centering
\footnotesize
\modernTableSetup
\setlength{\extrarowheight}{2pt}
\renewcommand{\arraystretch}{1.16}
\setlength{\tabcolsep}{1.5pt}
\caption{\textbf{Real-robot success rate (\%) by task horizon.} The \planner{} backbone is fixed across all methods; rows vary the agent-layer orchestration and navigation executor. Each tier contains 20 episodes.}
\label{tab:real_robot}
\renewcommand{\tabularxcolumn}[1]{m{#1}}
\begin{tabularx}{\columnwidth}{
    >{\raggedright\arraybackslash}p{0.17\columnwidth}
    >{\raggedright\arraybackslash}X
    S[table-format=2.0,table-column-width=0.115\columnwidth]
    S[table-format=2.0,table-column-width=0.115\columnwidth]
    S[table-format=2.0,table-column-width=0.115\columnwidth]
    S[table-format=2.1,table-column-width=0.115\columnwidth]}
\toprule
\rowcolor{TableHeader}
& & \multicolumn{1}{c}{\textbf{Low}}
& \multicolumn{1}{c}{\textbf{Medium}}
& \multicolumn{1}{c}{\textbf{High}}
& \\
\rowcolor{TableHeader}
\multirow{-2}{*}{\shortstack[l]{\textbf{Agent}\\\textbf{layer}}}
& \multirow{-2}{*}{\shortstack[l]{\textbf{Executor}\\\textbf{layer}}}
& \multicolumn{1}{c}{{\scriptsize single-room}}
& \multicolumn{1}{c}{{\scriptsize cross-room}}
& \multicolumn{1}{c}{{\scriptsize over 20\,m}}
& \multicolumn{1}{c}{\multirow{-2}{*}{\textbf{Overall}}} \\
\midrule
\rule{0pt}{2.2ex}Reactive & \executor{} & 80 & 35 & 0 & 38.3 \\
\system{} & Frontier exploration & 70 & 60 & 15 & 48.3 \\
\rowcolor{TableHighlight}
\system{} & \executor{}
& \multicolumn{1}{c}{\textbf{90}}
& \multicolumn{1}{c}{\textbf{85}}
& \multicolumn{1}{c}{\textbf{60}}
& \multicolumn{1}{c}{\textbf{78.3}} \\
\bottomrule
\end{tabularx}
\end{table}

\paragraph{Real-world transfer.}
\Cref{tab:real_robot} shows that \system{} reaches 78.3\% overall, versus 48.3\% for frontier exploration and 38.3\% for the reactive learned-navigator baseline.
The same interface therefore remains effective under real sensing and execution noise without privileged information.

\paragraph{Effect of exploration horizon.}
Margins over the strongest baseline rise from 10 to 25 and 45 points across tiers, demonstrating larger gains at longer horizons.
See \Cref{sec:appendix_real_robot,fig:deployment_vl_agent} for details.

\section{Conclusion, Limitations, and Future Work}
\label{sec:conclusion}

We presented \system{}, an agentic scaffolding framework that combines VLM reasoning with grounded NFM execution by closing the \gapname{}: the mismatch between route instructions and evidence-seeking intent, intermediate observation loss, and the absence of cross-call accumulation.
The navigator acts not as a standalone answerer but as an evidence-acquisition executor controlled by a VLM through intent, observation, and memory channels.
By separating evidence planning from execution, converting trajectories into reusable journey summaries, and preserving grounded observations across turns, \system{} improves three EQA benchmarks and controlled ablations, including a matched terminal-only interface baseline that isolates the cost of the gap itself.
Thus, navigation models benefit EQA most when an outer agent can inspect, reuse, and control their rollouts.

\paragraph{Limitations.}
\label{sec:limitations}

\system{} relies on a large VLM, slowing inference; smaller agents or distillation may reduce this cost.
Cross-view duplication also impairs counting.

\paragraph{Future Work.}
Beyond EQA, agents could use navigation foundation models to explore physical environments and efficiently collect embodied data for training agents' spatial intelligence.
Such agent-directed exploration may provide scalable, task-driven data that couples high-level reasoning with grounded observations and navigation trajectories.

\clearpage
\bibliographystyle{plainnat}
\bibliography{references}

\clearpage
\appendix

\section{Additional Experimental Analysis}
\label{sec:appendix_experimental_analysis}

\subsection{Per-Category Results}
\label{sec:per_category}

\begin{table}[H]
\centering
\small
\modernTableSetup
\renewcommand{\arraystretch}{1.16}
\caption{\textbf{Per-category results on HM-EQA.}}
\label{tab:per_category}
\begin{tabularx}{\columnwidth}{X S[table-format=2.1,table-column-width=0.30\columnwidth]}
\toprule
\rowcolor{TableHeader}
\textbf{Category} & {\textbf{Accuracy (\%)}} \\
\midrule
Existence & 80.6 \\
Identification & 80.2 \\
Location & 78.5 \\
State & 77.2 \\
Count & 65.0 \\
\midrule
\rowcolor{TableHighlight}
\textbf{Overall} & \multicolumn{1}{c}{\textbf{76.7}} \\
\bottomrule
\end{tabularx}
\end{table}

\Cref{tab:per_category} shows per-category performance.
Existence and identification questions achieve over 80\%, indicating that \system{} is effective when the answer can be grounded in object presence or category recognition after active search.
State questions also remain strong at 77.2\%, suggesting that the harness benefits verification-oriented questions where the agent must inspect candidate evidence rather than answer from a single observation.
Counting remains the hardest category at 65.0\%, partly because multiple keyframes may contain overlapping views of the same object and the agent can over-count repeated observations.
This result motivates stronger cross-view object association and uncertainty tracking for future versions.
On MT-HM3D, relationship questions reach 63.9\% and comparison questions reach 59.6\%.
Relationship questions benefit from spatial perception tools, while comparison questions benefit from the protocol requirement that target observations be recorded with source evidence before later reasoning turns.

\subsection{Detailed Ablation Analysis}
\label{sec:appendix_ablation_analysis}

The executor-capability block in \Cref{tab:capability_sides} shows a layered effect.
Replacing heuristic movement with StreamVLN improves accuracy from 60.9\% for Random Walk and 65.3\% for Frontier Exploration to 69.3\%, confirming the value of language-conditioned learned navigation.
Using \executor{}-4B further raises accuracy to 73.3\%, 4.0 percentage points above StreamVLN, while scaling to 8B adds a comparatively modest 0.7 percentage points.
Because these estimates use three stochastic hosted-LLM runs, we report the 4B--8B difference only as a scaling trend.

\paragraph{Episodic-interface baseline.}
The \emph{Episodic interface} row of \Cref{tab:ablation_internal} isolates the \gapname{} as a whole rather than one channel.
This variant keeps the agent and executor backbones, perception tools, episodes, initial states, and budgets identical to the full system, but replaces the three-channel protocol with the episodic tool interface defined in \Cref{sec:intro}. The agent issues a navigation instruction and receives the success flag, stop reason, executed step count, final pose, and final front view, without the structured intent schema, sampled trajectory keyframes, journey artifact, or protocol-managed cross-call evidence state. It does not alter the navigation model's internal policy or observation stream during execution.
The agent may still gather evidence at the arrival location through panoramic inspection, so the variant measures what the interface discards, not a crippled agent.
Accuracy falls to \termAcc\% ($-$\termDelta{} percentage points), a substantially larger drop than any single-component ablation.
This result shows that the whole interface gap is not captured by weakening journey analysis or persistent evidence state in isolation.

\paragraph{Terminal-only observation return.}
The observation-channel row at \obsTermAcc\% keeps the full intent and memory channels but changes only the navigation return to terminal status and the final front view.
Its \obsTermDelta{}-point drop isolates the cost of discarding en-route observations while retaining structured intent and cross-call accumulation.

\Cref{tab:ablation_internal} provides the detailed protocol ablations summarized in the main text.
Frontier Exploration exceeds Random Walk by 4.4 percentage points in \Cref{tab:capability_sides}, but both remain substantially below the learned navigation executors.
This gap is consistent with geometric coverage being imperfectly aligned with question-relevant evidence when the agent repeatedly issues semantic sub-goals; the largest drops occur on counting and identification questions, where semantically guided viewpoint selection is especially important for localizing relevant objects.
Changing only the observation return to terminal status and the final view reduces accuracy by \obsTermDelta{} percentage points, directly measuring the value of en-route observations.
Removing journey analysis reduces accuracy by 4.4 percentage points because intermediate observations are no longer converted into a structured journey artifact for the agent.
Reducing keyframe density costs 2.8 percentage points, confirming that visual sampling during navigation directly affects the agent's environmental state.
Restricting the executor to a single navigation mode drops accuracy by 2.0 percentage points, indicating that object-targeted and instruction-following calls support complementary exploration strategies.
Context compaction has a smaller direct accuracy effect (0.9 percentage points), which suggests that its main role is to keep long-horizon context management efficient and stable rather than to act as a stand-alone accuracy module.
Removing \texttt{zoom\_in\_object} costs 0.6 percentage points, indicating that local verification is useful but not the primary driver of the harness gains.
Among the remaining component ablations, removing the EQA context state causes the largest drop, from 74.0\% to 69.4\% ($-$4.6 percentage points).
This result shows that the agent cannot rely on raw turn-by-turn observations alone: it needs an explicit EQA context state that records accumulated evidence in the evidence ledger, unresolved goals in the unresolved-goal state, and answer-relevant history across navigation calls.
Without this state, multi-turn exploration becomes less coherent, and evidence collected in earlier trajectories is less reliably reused during final answer verification.

\section{Evaluation Protocol and Statistics}
\label{sec:appendix_statistics}

Baseline values in \Cref{tab:main} are quoted from the corresponding papers; $^*$ denotes results reproduced and reported by FAST-EQA.
Because several implementations and per-episode predictions are unavailable, cross-paper differences are contextual only and are excluded from paired significance tests.
Our main results are mean $\pm$ standard deviation over three independent runs, where the variation arises from the stochasticity of the hosted LLM API rather than from explicit random seeds.
For the matched-task comparison in \Cref{tab:matched_systems}, Explore-EQA retains its own pipeline, while ToolEQA and FAST-EQA are independently reproduced by us following their reported methods. All three use the same Qwen3.5 reasoning model, benchmark episodes, initial states, and interaction budget as the \system{} reference in that table.
In addition, the ToolEQA reproduction is equipped with the same perception and verification toolset as \system{} (\Cref{tab:tools}), covering panoramic inspection, open-vocabulary detection and segmentation, close-up object inspection, depth estimation, and saved-keyframe review, backed by the same perception service and VLM backends.
ToolEQA invokes its original episodic navigation tools within its own pipeline, whereas \system{} issues semantic navigation calls and receives trajectory-level evidence through its protocol channels.
All executor variants share the agent, perception stack, split, episode initialization, and navigation budget; only the executor changes. We report these controlled results over three independent runs.
The episodic-interface variant in \Cref{tab:ablation_internal} holds the agent and executor backbones, perception stack, episodes, and budgets fixed while replacing the three-channel protocol; the terminal-only observation-return variant changes only the navigation return format. Both are evaluated across three runs.

\section{Real-Robot Evaluation Details}
\label{sec:appendix_real_robot}

Each episode starts from a natural-language query, resets the exploration state, and allows the agent to issue multiple navigation calls.
The robot records egocentric RGB observations, pose estimates, an online explored map, keyframes, searched regions, negative observations, and final answer support, without privileged object locations or unseen geometry.
The two \emph{low} tasks require single-room evidence gathering (e.g., the logo printed on a backpack or the color of a water dispenser), the two \emph{medium} tasks require cross-room search (e.g., the screen-cast code in meeting room B01 or the nearest coffee-shop brand), and the two \emph{high} tasks require open-scene exploration over routes exceeding 20\,m (e.g., the number of elevators in the C3 elevator hall or whether the post office is open).

\paragraph{Episode construction.}
For each task, we predefine two distinct robot starting poses and repeat each starting condition five times, producing $2\times5=10$ evaluation episodes.
The two tasks in each difficulty tier yield 20 episodes per method and tier, for a total of 60 real-robot episodes per method.
All compared methods are evaluated on exactly the same starting poses, natural-language queries, and environment configuration.
Before each episode, the robot is returned to the designated starting pose, and the exploration state and agent memory are reset.

\paragraph{Success and safety protocol.}
Episode success is judged manually after the run.
An episode is successful only if the robot autonomously completes the evidence-gathering process within the tier-specific time limit and returns the correct answer supported by observations acquired during that episode.
The timeout is 5 minutes for low-difficulty tasks, 10 minutes for medium-difficulty tasks, and 15 minutes for high-difficulty tasks; a timeout or an incorrect answer is counted as failure.
The Unitree Go2 local obstacle-avoidance system remains enabled throughout every run to ensure safe physical execution.
The low-level safety layer supplies no semantic targets, answer evidence, or planning.

Because physical runs are costly, this study targets sim-to-real transfer rather than repeating the full capability and protocol ablations established in \Cref{tab:capability_sides,tab:ablation_internal}.
The reactive baseline keeps the learned navigator but removes multi-step deliberation, isolating memory-driven orchestration between the agent and navigator; the frontier baseline replaces the learned navigator with heuristic coverage, testing query-conditioned semantic navigation under real odometry drift and local recovery.

On single-room tasks all methods perform reasonably (70--90\%), because the answer lies within one room.
On cross-room tasks the reactive variant drops to 35\% without persistent deliberation, whereas \system{} reaches 85\%.
On long-horizon tasks the reactive variant fails (0\%) and frontier coverage reaches 15\%, while \system{} answers 60\% of episodes.
The widening gap mirrors simulation: longer, multi-room tasks require persistent evidence and multi-step orchestration beyond a navigator or single-shot reasoner.

\section{Qualitative Analysis}

\begin{figure}[H]
\centering
\includegraphics[width=0.82\textwidth]{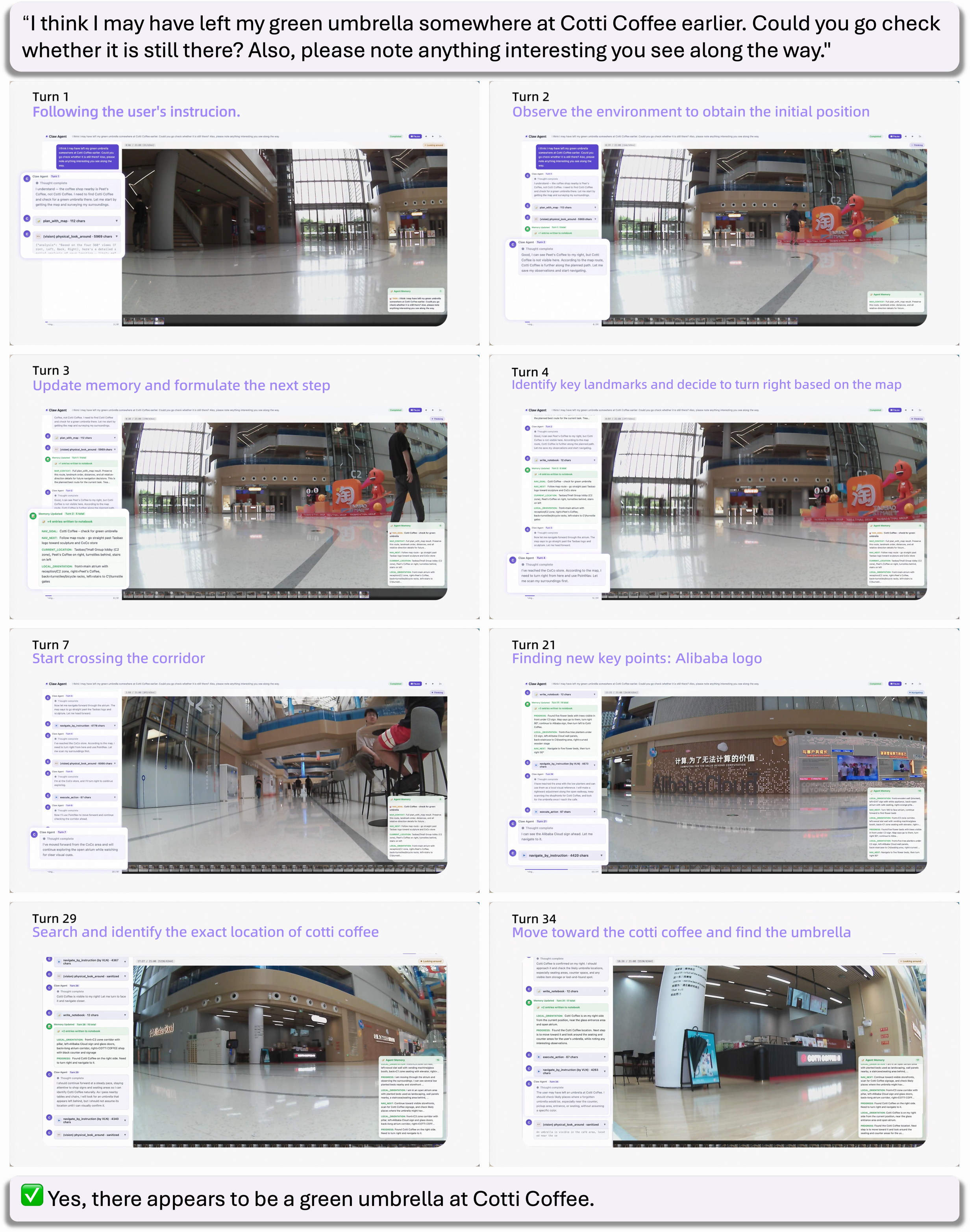}
\caption{\textbf{Ultra-long-horizon real-robot navigation with agentic \executor{}.}
On a physical robot, the agent answers an open-ended request over a trajectory longer than $50\,\mathrm{m}$ by decomposing the task into sub-goals, following landmarks to Cotti Coffee, and verifying the green umbrella from visual evidence.
Selected turns illustrate the loop of upper-level planning, \executor{} execution, memory updates, and final response generation.}
\label{fig:deployment_vl_agent}
\end{figure}

\begin{figure}[H]
\centering
\includegraphics[width=0.92\textwidth]{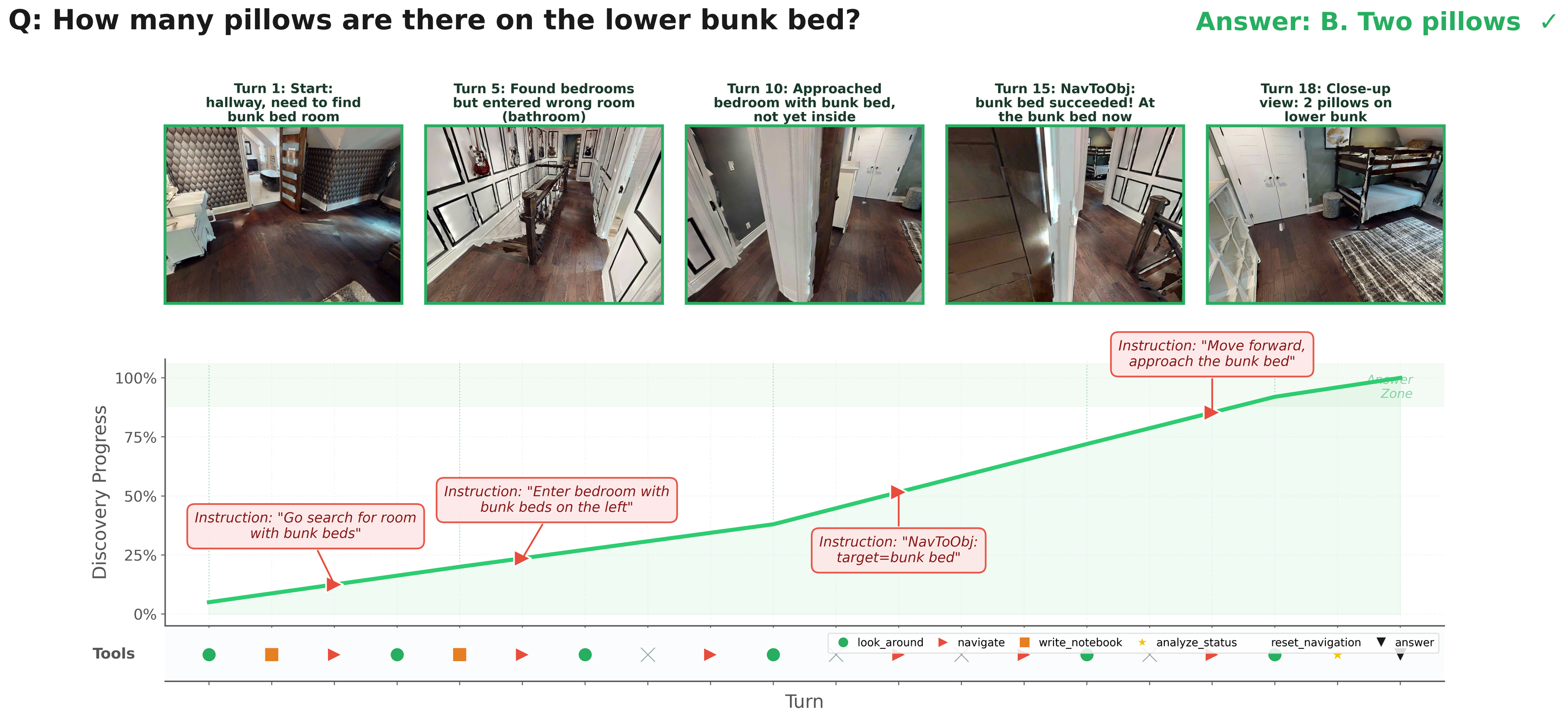}
\caption{\textbf{Additional progressive discovery episode.} The agent is asked to count pillows on a lower bunk bed. It first uses instruction-guided navigation to search for a room with bunk beds, switches to \texttt{navigate\_to\_object(target=bunk bed)} after reaching the bedroom area, and then uses a final local approach instruction to obtain a closer view. This appendix case complements \Cref{fig:progressive_discovery} by showing how the same protocol also supports fine-grained answer verification after object localization.}
\label{fig:appendix_progressive_discovery}
\end{figure}

\paragraph{Ultra-long-horizon real-robot showcase.}
\Cref{fig:deployment_vl_agent} presents a physical deployment whose cumulative navigation trajectory exceeds $50\,\mathrm{m}$, substantially extending beyond the $>20\,\mathrm{m}$ open-scene tier evaluated in the main real-robot study.
Rather than attempting the request in a single navigation call, the agent decomposes it into landmark-conditioned sub-goals, uses Cotti Coffee as an intermediate anchor, records progress in memory, and continues until the requested green umbrella is visually verified.
This episode shows that the protocol between the agent and navigator preserves task state and grounded evidence over an ultra-long physical trajectory and repeated planning and execution cycles.

We qualitatively inspect successful episodes to understand how the harness uses navigation calls as evidence-gathering steps rather than as isolated movements.
In multi-room search questions, the agent first issues broad instructions such as searching nearby bedrooms or kitchen areas, then records which regions have been checked before selecting the next sub-goal.
This behavior reduces repeated exploration because later turns can condition on notebook entries such as ``Kitchen searched: microwave and toaster found; coffee maker absent'' instead of relying only on the latest view.

The traces also show adaptive switching between navigation modes.
When the target category is still uncertain, the agent tends to call \texttt{navigate\_by\_instruction} with a coarse semantic direction, allowing the executor to search for a plausible area.
After an object anchor is observed, the agent often switches to \texttt{navigate\_to\_object} or visual verification tools to obtain closer evidence before answering.
This pattern matches the intent-channel design: the agent specifies the evidence need at the semantic level, while the executor handles low-level movement.
The additional episode in \Cref{fig:appendix_progressive_discovery} shows this second pattern: once the agent narrows the search to the bedroom area, an object-targeted call and a short local approach provide the visual evidence needed to count the pillows on the lower bunk bed.

Failure cases are concentrated in questions that require precise counting or fine-grained state verification.
In these episodes, multiple keyframes may contain overlapping views of the same object, and the agent sometimes treats repeated observations as distinct evidence.
Future versions should improve cross-view object association and uncertainty tracking, especially for counting, the weakest category.

\clearpage
\begin{table}[H]
\centering
\small
\modernTableSetup
\renewcommand{\arraystretch}{1.16}
\setlength{\tabcolsep}{3pt}
\caption{\textbf{Real-robot evaluation protocol.} Each tier contains two tasks. Two fixed starting poses with five repeated episodes per pose produce ten episodes per task and twenty episodes per method and tier.}
\label{tab:real_robot_protocol}
\begin{tabularx}{\columnwidth}{X c c c c}
\toprule
\rowcolor{TableHeader}
\textbf{Tier} & \textbf{Starts} & \textbf{Runs/start} & \textbf{Episodes} & \textbf{Timeout} \\
\midrule
Low & 2 & 5 & 20 & 5 min \\
Medium & 2 & 5 & 20 & 10 min \\
High & 2 & 5 & 20 & 15 min \\
\bottomrule
\end{tabularx}
\end{table}

\FloatBarrier
\section{Tool Specification}
\label{sec:appendix_tools}

\Cref{tab:tools} provides the complete tool specification exposed to the VLM agent via the tool-calling API.
Exposure denotes availability in the agent's tool schema; it does not imply that a tool is selected in every episode.
The optional cross-episode retrieval store is disabled in all reported experiments and is therefore excluded from the table.

\begin{table}[H]
\centering
\small
\modernTableSetup
\renewcommand{\arraystretch}{1.14}
\caption{\textbf{Complete tool specification} exposed to the agent through the tool-calling interface.}
\label{tab:tools}
\begin{tabularx}{\textwidth}{p{0.13\textwidth} p{0.27\textwidth} X}
\toprule
\rowcolor{TableHeader}
\textbf{Category} & \textbf{Tool} & \textbf{Description} \\
\midrule
\multirow{8}{*}{Visual} & \texttt{look\_around} & Read four directional views with the VLM for scene context and navigation affordances \\
& \texttt{analyze\_status} & Arbitrate current, historical, and notebook evidence before answer commitment \\
& \texttt{detect\_objects\_360} & Detect and count prompted objects across all four views \\
& \texttt{zoom\_in\_object} & Crop a detected object and inspect fine-grained attributes with full-view context \\
& \texttt{review\_image} & Retrieve and re-analyze archived panoramas or navigation keyframes \\
& \texttt{segment} & Segment a text-prompted object in the current front view \\
& \texttt{detect\_objects} & Run prompted or open-vocabulary detection in the current front view \\
& \texttt{estimate\_depth} & Estimate metric depth and object-contact information in the current front view \\
\midrule
\multirow{4}{*}{Navigation} & \texttt{navigate\_to\_object} & Object-goal navigation (up to $S$ steps) \\
& \texttt{navigate\_by\_instruction} & Instruction-following navigation (up to $S$ steps) \\
& \texttt{reset\_navigation} & Clear executor-side history before switching to a new destination \\
& \texttt{execute\_action} & Apply one local turn, move, or stop action for final-view alignment \\
\midrule
Context & \texttt{write\_notebook} & Append observations to the persistent within-episode evidence notebook \\
\bottomrule
\end{tabularx}
\end{table}

\FloatBarrier
\subsection{Perception and Verification Tools}
\label{sec:appendix_perception_tools}

The perception tools use two complementary backends.
Reader tools (\texttt{look\_around}, \texttt{analyze\_status}, \texttt{review\_image}, and \texttt{zoom\_in\_object}) send visual evidence to the VLM with a task-specific prompt.
Detector tools (\texttt{segment}, \texttt{detect\_objects}, \texttt{detect\_objects\_360}, and \texttt{estimate\_depth}) call a separate perception service built on open-vocabulary segmentation and detection together with metric depth estimation.
Single-frame tools automatically use the current front view.
Panoramic tools use the four native front, right, back, and left cameras when available; otherwise the adapter synthesizes the panorama by rotating the agent through four $90^{\circ}$ headings and restoring its original orientation.

\paragraph{Panoramic scene reading and detection.}
\texttt{look\_around} sends the four direction-labeled views in one multi-image request and returns a free-text scene analysis together with the saved-image manifest.
Every reader prompt additionally asks the VLM to report visible doorways, corridors, stairs, open passages, and room transitions because these observations support the next navigation decision.
For light-state questions, the prompt explicitly distinguishes artificial illumination from sunlight.
\texttt{detect\_objects\_360} runs prompted open-vocabulary detection on all four views and aggregates per-label instances and the views in which they occur.
Because adjacent views can contain the same object, its return includes a warning that per-view counts require cross-view reconciliation.

\begin{minipage}{\linewidth}
\noindent\textit{Shared navigation-focus suffix.}
\begin{lstlisting}[style=promptstyle]
IMPORTANT: In addition to answering the question, always note:
doorways, hallways, corridors, staircases, open passages, and room
transitions visible in the image. These are critical for navigation
planning.

For light/lamp on-off questions: distinguish between NATURAL light
(sunlight from windows) and ARTIFICIAL light (lamp glow, light fixture
illumination). A lamp is ON only if it emits visible glow or
illumination. A bright room from window sunlight does NOT mean lamps
are on.
\end{lstlisting}
\end{minipage}

\begin{minipage}{\linewidth}
\noindent\textit{Panoramic scene-reading prompt.}
\begin{lstlisting}[style=promptstyle]
You are seeing 4 views (front, right, back, left) forming a 360 degree
panorama of your current location.
{question}{navigation_focus_suffix}
\end{lstlisting}
\end{minipage}

\paragraph{Fine-grained and historical inspection.}
\texttt{zoom\_in\_object} detects the requested target across the panorama, retains the largest-area detection, and sends both a padded crop and its full source view to the VLM.
The full view establishes global category and scene context, while the crop is used for color, pattern, subtype, and text.
When the two disagree, the full-view judgment takes precedence so that a context-free crop cannot overturn a reliable scene-level interpretation.
\texttt{review\_image} re-runs the VLM on images already stored during the episode, selected by path, turn, tool, view, panorama, or the latest navigation-keyframe shortcut.
It recovers evidence from intermediate observations without retraversing the region.

\begin{minipage}{\linewidth}
\noindent\textit{Fine-grained crop-and-context prompt.}
\begin{lstlisting}[style=promptstyle]
The first image is a cropped close-up of '{target}'. The second image
is the full {view} view for context.

Question: {question}

Focus on fine-grained visual attributes: exact color, pattern,
texture, shape, object type/subtype, brand, text/labels on the object.
Be precise and specific. For example, say 'dark green' not just 'green',
'induction cooktop' not just 'stove'.
\end{lstlisting}
\end{minipage}

\paragraph{Single-frame geometric tools.}
\texttt{segment} returns whether a text-prompted mask is found in the current front frame; \texttt{detect\_objects} lists prompted or open-vocabulary detections; and \texttt{estimate\_depth} combines segmentation with metric depth to report target distance and contact information.
These tools remain available, although panoramic inspection usually yields more complete EQA evidence.

\FloatBarrier
\section{Implementation Details}
\label{sec:appendix_impl}

\paragraph{Navigation executor.}
We use the pretrained 8B \executor{} checkpoint~\citep{zhang2026qwenrobotnav} as a fixed embodied executor. Given a natural-language sub-goal and egocentric observations, it predicts a short-horizon trajectory of eight planar waypoints. The adapter exposes only the two modes used in our EQA experiments: \texttt{navigate\_by\_instruction} for language-guided movement and \texttt{navigate\_to\_object} for object search and approach. Each call returns an executable waypoint rollout; answer-level reasoning, trajectory summarization, evidence-ledger updates, and the decision to issue another call remain responsibilities of \system{}. We do not train or modify the executor; architecture, training, and navigation-only evaluations are documented in the source report.

\paragraph{Closed-loop navigation adapter.}
Both semantic navigation tools invoke the same multi-step controller and differ only in how the language sub-goal is formed.
At each simulator step, the adapter sends the executor the current front, right, back, and left RGB observations, the natural-language instruction, episode identifier, agent position and orientation, online explored map, and collision state.
The executor returns either a stop decision or a relative short-horizon waypoint prediction.
The adapter transforms the prediction into a navmesh-aware waypoint action and executes it before acquiring the next observation.
The loop terminates when the executor stops, the simulator marks the episode complete, or the call exhausts its step budget $S=64$.
Two consecutive collisions set an \texttt{is\_stuck} flag in the next executor query, enabling closed-loop recovery rather than terminating the semantic call immediately.
\texttt{reset\_navigation} clears the executor's server-side episode history before the agent switches to a new destination, whereas \texttt{execute\_action} bypasses the executor for one local alignment action such as turning the final target into the front view.

\paragraph{Trajectory return.}
During a rollout, the adapter samples the front view every four simulator steps and retains at most 16 evenly spaced keyframes.
After execution, the trajectory summarizer converts these frames into the journey artifact described in \Cref{sec:appendix_protocol_details}.
Each navigation call returns the number of executed steps, executor stop state, simulator completion state, path information, journey analysis, and persisted image references.
This return exposes the navigation outcome and pre-terminal evidence.

\paragraph{Deployment.}
The main results use \planner{} as the VLM agent with a 256K token context window and native tool-calling support.
For controlled ablations, we use our self-deployed open-source Qwen3.5-397B-A17B agent.
Across ablation variants, we keep the agent backbone, serving stack, decoding configuration, and tool-calling setting fixed.
All variants use the same hosted 8B executor unless the executor itself is ablated in \Cref{tab:capability_sides}.
Visual perception uses an open-vocabulary detector with SAM for segmentation and Depth-Anything for depth estimation.

\paragraph{Computing infrastructure.}
Simulation and local perception (open-vocabulary detection, SAM segmentation, and Depth-Anything depth) run in Habitat-Sim 0.3.3 on HM3D scenes on one NVIDIA RTX 4090 GPU under Ubuntu 22.04. Because the \planner{} agent and \executor{} use hosted APIs on uncontrolled hardware, run-to-run variation reflects API stochasticity, not local nondeterminism.

\paragraph{Hyperparameters.}
Key \system{} settings: 50 agent turns; $S=64$ steps/call; $T=60{,}000$-token compaction; $k=2$ protected VLM results; 100 ledger entries; four-step sampling with 16 keyframes/call; 1024/2048 assessment/synthesis tokens; and a 4,000-character tool-result cap.

\FloatBarrier
\section{Protocol Details}
\label{sec:appendix_protocol_details}

\begin{algorithm}[H]
\small
\caption{Protocol between the agent and executor for long-horizon EQA}
\label{alg:agent_loop}
\begin{algorithmic}[1]
\STATE Initialize EQA context state $C_0 \leftarrow (H_0,E_0,U_0)$ from question $q$ and initial observation
\FOR{$t = 0$ to $T_{\max}-1$}
\STATE Agent reads $C_t$ and emits either a candidate answer $y_t$ or an evidence need $u_t$
    \IF{agent emits candidate answer $y_t$}
        \IF{$y_t$ is supported by the current observation, $E_t$, or source artifacts}
            \RETURN $y_t$
        \ENDIF
        \STATE Update $U_t$ with the missing evidence
        \STATE \textbf{continue}
    \ENDIF
    \STATE \textsc{Intent interface:} translate $u_t$ into navigation call $a_t=(m_t,g_t,S_t,c_t)$
    \STATE \textsc{Executor call:} run $a_t$ with the navigation foundation model and return trajectory record $r_t=(s_t,b_t,\ell_t,p_t,K_t)$
    \STATE \textsc{Observation interface:} convert $r_t$ into source-grounded journey evidence $z_t=(q_t,R_t,O_t,P_t,D_t)$
    \STATE \textsc{Memory interface:} update $E_{t+1}$ and $U_{t+1}$ from $(C_t,z_t)$ with source references
    \STATE Compact processed raw returns and set $C_{t+1}=(H_{t+1},E_{t+1},U_{t+1})$
\ENDFOR
\STATE Return the best answer supported by $C_{T_{\max}}$
\end{algorithmic}
\end{algorithm}

\paragraph{Intent Call Schema.}
At turn $t$, the agent emits
\[
a_t = (m_t, g_t, S_t, c_t),
\]
where $m_t$ is the navigation mode, $g_t$ is a natural-language sub-goal, $S_t$ is the step budget, and $c_t$ contains optional constraints such as target attributes or preferred regions.
At each simulation step, the executor observes four RGB views, its current pose, and an online occupancy map built only from the explored trajectory; unexplored geometry and object locations remain unavailable.
Execution stops when the executor emits \textsc{stop}, exhausts $S_t$, or the simulator marks the episode complete.
Repeated collisions instead trigger recovery by setting \texttt{is\_stuck} in the next executor query; they do not immediately terminate the call.

\paragraph{Trajectory Record Schema.}
For a navigation call $a_t$, the executor first returns a raw trajectory record
\[
r_t = (s_t, b_t, \ell_t, p_t, K_t),
\]
where $s_t$ is the success flag, $b_t$ is the stop reason, $\ell_t$ is the path length, $p_t$ is the final pose, and $K_t=\{I_{t,1},\ldots,I_{t,n}\}$ is a bounded set of sampled keyframes.

\paragraph{Keyframe Sampling.}
For each navigation call, keyframes are sampled every $\Delta=4$ steps and capped at $M=16$ frames per call.
This setting balances visual coverage against context cost, while still allowing the observation channel to preserve evidence that appears before the executor stops.

\paragraph{Journey Artifact Schema.}
A VLM summarizer outputs
\[
z_t = (q_t, R_t, O_t, P_t, D_t).
\]
Here $q_t$ records the attempted sub-goal and whether it appeared to be satisfied.
$R_t$ summarizes rooms and route-level context, and $O_t$ stores salient observed objects.
$P_t$ stores planning cues such as doors, corridors, and exits, while $D_t$ records uncertainty or failure conditions.

\paragraph{Source-Grounded Journey Artifact.}
Each object or room mention in the journey artifact is tied to a source keyframe and confidence label.
We represent a grounded mention as $(n,i,v,h,c)$, where $n$ is the name, $i$ is the keyframe id, $v$ is the viewpoint, $h$ is a location hint, and $c$ is the confidence label.
The summarizer is instructed to report only visually observed evidence, mark uncertainty explicitly, and avoid turning absence from a single view into a definitive negative claim.
The prompt template used to narrate the sampled trajectory is shown below; the task-question line is included only when an EQA question is available.

\noindent\textit{Journey-analysis prompt.}
\begin{lstlisting}[style=promptstyle]
The agent was navigating with instruction: "{instruction}"
The agent's task question is: "{question}"
These images show the agent's front view at different points during
navigation.
For each step, briefly describe: what room/area is this? What key
objects, doorways, hallways, stairs, or room transitions are visible?
If any details relevant to the task question are directly visible,
mention them.
Only mention question-relevant details when they are directly visible
in the images; do not infer unseen states.
Keep each step description to 1-2 sentences.
\end{lstlisting}

\paragraph{Evidence arbitration with \texttt{analyze\_status}.}
\label{sec:appendix_analyze_status}

\texttt{analyze\_status} is the terminal evidence-verification tool and operates in three phases.
First, the VLM jointly reads the current four views, exploration notebook, and available archived visual evidence and produces a structured assessment of target visibility and evidence sufficiency.
Second, when evidence is insufficient, the harness may dispatch one supplementary inspection (saved-keyframe review for historical detail or panoramic detection for object presence and counting) and return its result to the same reasoning context.
Third, the VLM synthesizes a final status that either authorizes an answer, records well-supported negative evidence, or directs further exploration.
The selected \texttt{target\_view} also tells the agent whether a local rotation is needed to place the target in the final front view.

\noindent\textit{Phase 1: structured assessment prompt.}
\begin{lstlisting}[style=promptstyle]
## Exploration Notebook
{notebook_text}

## Analysis Question
{question}

You have the current 360 degree views, retrieved visual memory, and the
agent's exploration notebook.
The scene is static. All object states remain unchanged throughout the
episode.

When memory images are provided, explicitly compare [MEMORY] evidence
against [CURRENT] views before deciding whether evidence is sufficient.

Based on ALL evidence above, respond with a JSON object:
{
  "target_visibility": "clear" | "partial" | "not_seen" | "uncertain",
  "target_view": "front" | "right" | "back" | "left" | null,
  "enough_evidence": true/false,
  "provisional_answer": "brief answer or null",
  "reasoning_summary": "1-2 sentence reasoning",
  "missing_evidence_type": "none" | "historical_detail" |
                           "object_presence" | "counting" |
                           "fine_grained_attribute",
  "recommended_tool": "none" | "review_image" |
                      "detect_objects_360" | "zoom_in_object",
  "tool_target": "object name for the tool, or null"
}
\end{lstlisting}

The final synthesis prompt imposes a deliberately asymmetric evidence rule: a positive answer requires visible support, while a negative answer requires coverage of the relevant area rather than mere failure to observe the target.

\noindent\textit{Phase 3: final synthesis prompt.}
\begin{lstlisting}[style=promptstyle]
## Final Synthesis
Question: {question}

Based on the 360 degree views, notebook, and any supplementary evidence
above, provide your FINAL ANALYSIS.

First, state your STATUS as one of:
- ANALYSIS_COMPLETE: You have enough evidence to confidently answer.
- NEGATIVE_EVIDENCE: You can confidently conclude the target does NOT
  exist or the condition is NOT met. You may ONLY use this status when:
  (a) The current 360 degree views clearly cover the relevant area and
      the target is obviously absent, OR
  (b) Historical navigation keyframes + current views + notebook
      consistently confirm the relevant area has been thoroughly
      checked.
  Do NOT use NEGATIVE_EVIDENCE if you simply haven't seen the target
  yet. That is INSUFFICIENT_EVIDENCE.
- INSUFFICIENT_EVIDENCE: You genuinely lack evidence because the area has not
  been sufficiently explored.

Format: Start with STATUS: <status> on the first line, then your
analysis. Be specific and cite what you see. Do not output JSON.
\end{lstlisting}

\paragraph{Evidence Ledger Schema.}
Each evidence entry is
\[
e_i = (\tau_i, \sigma_i, \kappa_i, \eta_i, \gamma_i, u_i),
\]
where $\tau_i$ is the turn index and $\sigma_i$ is the source artifact or keyframe.
$\kappa_i$ is the evidence claim, $\eta_i$ records its support, $\gamma_i$ is a confidence label, and $u_i$ marks whether the entry remains unresolved.
The unresolved-goal state tracks goals, uncertain findings, and regions needing inspection.

\paragraph{Notebook implementation.}
\texttt{write\_notebook} implements the persistent within-episode evidence ledger exposed to the agent.
It accepts a list of concise observations, tags each new entry with its source turn, removes normalized exact duplicates, and retains at most 100 entries in FIFO order.
The notebook is reconstructed in the agent prompt on every turn, so curated evidence remains available even after older raw tool returns are compacted.
Entries record navigation goals, visited or searched regions, room transitions, question-relevant positive and negative observations, uncertainty, and remaining evidence needs.
The notebook resets for every evaluation episode; the reported results use no cross-episode retrieval or test-environment memory.

\paragraph{Context Compression.}
Raw tool returns are compressed only after salient information has been externalized into the evidence ledger or unresolved-goal state with source references.
After this update, older tool returns are replaced with lightweight placeholders.
If the total context exceeds threshold $T$, \system{} keeps the system prompt, the original task input, the evidence ledger, unresolved goals, and recent dialogue turns, while dropping intermediate messages.

\paragraph{Equivalent Step Computation.}
\label{sec:appendix_steps}
Prior frontier-based EQA methods, including Explore-EQA, Memory-EQA, and FAST-EQA, define one exploration step as a high-level sense-plan-act cycle in which the agent teleports to a selected frontier point.
When the VLM directly controls exploration in these protocols, one selected frontier action displaces the agent by at most 3 meters.
We therefore inherit 3 meters as the conversion unit rather than choosing it as a free threshold, and the per-episode step budget is
\begin{equation}
N = \left\lfloor \sqrt{A} \times 3 \right\rfloor,
\end{equation}
where $A$ is the navigable floor area of the scene in square meters.
The normalized step metric is then
\begin{equation}
\hat{n} = \frac{n_{\text{actual}}}{N},
\end{equation}
which measures the fraction of the allowed exploration budget consumed by an episode.

For \system{}, let $H$ be exactly the number of outer-agent VLM invocations during an episode.
Each invocation is charged one base step, regardless of the high-level decision it returns.
However, one semantic navigation decision may trigger a continuous VLN rollout longer than the 3-meter movement allowed to a frontier method in one step.
Counting such a rollout as only one step would favor \system{}.
We therefore add a distance compensation for every NFM navigation call.
Let $K$ be the number of NFM navigation calls, and let $L_k$ be the cumulative executed path length of the $k$-th NFM rollout, computed from its trajectory waypoints.
The added charge is
\begin{equation}
c_k = \max\!\left(0,\left\lceil\frac{L_k}{3\,\mathrm{m}}\right\rceil-1\right).
\end{equation}
The compensated metrics are
\begin{equation}
n_{\text{eq}} = H + \sum_{k=1}^{K} c_k,
\qquad
\hat{n}_{\text{eq}} = \frac{n_{\text{eq}}}{N}.
\end{equation}
Thus, a rollout of at most 3 meters receives no extra charge, while rollouts of $(3,6]$ and $(6,9]$ meters receive one and two additional steps, respectively.
The ceiling operation deliberately favors the frontier baselines: each additional partial 3-meter segment is charged as a full step, and $L_k$ includes detours, repeated motion, and collision recovery that frontier teleportation does not incur.
In our setup, outer-agent VLM reasoning and API latency dominate wall-clock time, so $H$ is the primary efficiency quantity.
This accounting is not a count of all model inference in the system.
Calls made by the navigation executor, trajectory summarizer, and auxiliary perception models are not separately charged; the outer-agent invocations that decide to call these components or reason over their outputs are already included in $H$.
Accordingly, $\hat{n}_{\mathrm{eq}}$ measures normalized high-level interaction and exploration rather than end-to-end computational cost.
This equivalent-step compensation is applied only to configurations that invoke an NFM; non-NFM methods retain their original direct-control step accounting.

\paragraph{Collision Recovery.}
The executor tracks consecutive collisions during navigation.
After 2 or more consecutive collisions, an \texttt{is\_stuck} signal is sent to the navigation foundation model, which can then adjust its predicted trajectory to recover from the stuck state.

\end{document}